\pdfoutput=1

\documentclass[11pt]{article}

\usepackage[]{acl}

\usepackage{times}
\usepackage{latexsym}
\usepackage{amsmath} 
\usepackage{graphicx}
\usepackage{amsfonts}
\DeclareMathOperator*{\argmax}{arg\,max}

\usepackage[T1]{fontenc}

\usepackage[utf8]{inputenc}

\usepackage{microtype}

\title{LMN at SemEval-2022 Task 11: A Transformer-based System for English Named Entity Recognition}

\author{Ngoc Minh Lai \\
  Van Ho Middle School \\
  \texttt{nl834771@gmail.com}}

\begin{document}
\maketitle
\begin{abstract}
Processing complex and ambiguous named entities is a challenging research problem, but it has not received sufficient attention from the natural language processing community. In this short paper, we present our participation in the English track of SemEval-2022 Task 11:  Multilingual Complex Named Entity Recognition. Inspired by the recent advances in pretrained Transformer language models, we propose a simple yet effective Transformer-based baseline for the task. Despite its simplicity, our proposed approach shows competitive results in the leaderboard as we ranked 12 over 30 teams. Our system achieved a macro F1 score of 72.50\% on the held-out test set. We have also explored a data augmentation approach using entity linking. While the approach does not improve the final performance, we also discuss it in this paper.
\end{abstract}

\section{Introduction}
Recognizing complex named entities (NEs) is a challenging research problem, but it has not received sufficient attention from the natural language processing community \cite{mengetal2021gemnet,fetahu2021gazetteer}. Complex NEs can be complex noun phrases (e.g., \textit{National Baseball Hall of Fame and Museum}), gerunds (e.g., \textit{Saving Private Ryan}), infinitives (e.g., \textit{To Build a Fire}), or even full clauses (e.g., \textit{I Capture The Castle}). This ambiguity makes it difficult to recognize them based on their context \cite{aguilaretal2017multi,lukenetal2018qed,hanselowskietal2018ukp}.

In this paper, we describe our participation in the English track of SemEval-2022 Task 11: Multilingual Complex Named Entity Recognition \cite{multiconerData,multiconerReport}.
Inspired by the recent success of Transformer-based pre-trained language models in many NLP tasks \cite{devlinetal2019bert,joshietal2019bert,lai2019gated,joshietal2020spanbert,tranetal2020explain,yuetal2020named,wen2021resin,lai2021joint,Monaikul2021ContinualLF}, we propose a simple but effective Transformer-based baseline for the task. Despite its simplicity, our proposed approach shows promising results: the official ranking indicated that our system achieved a macro $\text{F}_1$ score of 72.50\% on the test set and ranked 12th out of 30 teams. We have also explored a data augmentation approach using entity linking. While the approach does not improve the final performance, we also discuss it in this paper.

In the following sections, we first describe the related work in Section \ref{sec:related_work} and the proposed method in Section \ref{sec:method}. We then describe the experiments and their results in Section \ref{sec:results}. Finally, Section \ref{sec:conclusion} concludes this work and discusses potential future research directions.

\begin{figure*}[ht!]
\centering
\includegraphics[width=0.8\textwidth]{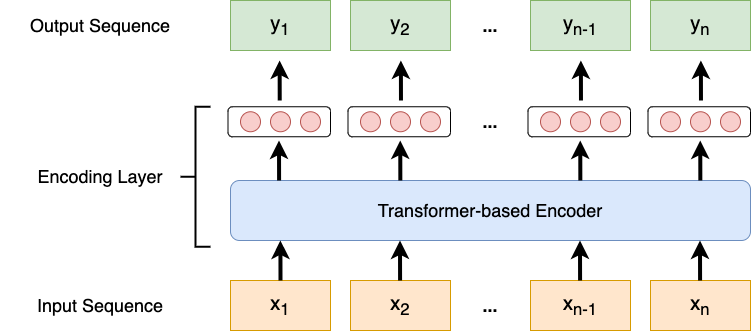}
\caption{Overview of our Transformer-based model.}
\label{fig:overview}
\end{figure*}

\section{Related Work} \label{sec:related_work}
Many previous named entity recognition (NER) methods are based on the sequence labeling approach \cite{Collobert2011NaturalLP,mahovy2016end,lampleetal2016neural,chiunichols2016named,biobert,yangetal2018design,yangzhang2018ncrf,lai2020joint,lietal2020handling}. For example, \newcite{Collobert2011NaturalLP} introduced a neural architecture that uses convolutional neural networks (CNNs) to encode tokens combined with a CRF layer for the classification. Many other studies used recurrent neural networks (RNNs) instead of CNNs to encode the input and a CRF for the prediction \cite{mahovy2016end,lampleetal2016neural}. With the recent rise of pre-trained language models, recent NER models typically make use of context-dependent embeddings such as ELMo \cite{Peters2018DeepCW} or BERT \cite{devlinetal2019bert}.

While neural-based models have achieved impressive results on popular benchmark datasets like CoNLL03 and OntoNotes \cite{tjongkimsang2002introduction,tjongkimsangdemeulder2003introduction,pradhanetal2012conll}, these models typically do not perform well on complex/unseen entities \cite{Augenstein2017GeneralisationIN}. Complex named entities (e.g., titles of creative works) are typically not simple nouns and are harder to recognize. The challenges of NER for recognizing complex entities and in low-context situations was recently outlined by \citet{meng2021gemnet}. Other work has extended this to multilingual and code-mixed settings \cite{fetahu2021gazetteer}.

\section{Method} \label{sec:method}
\subsection{Baseline model}
Similar to many previous studies \cite{lampleetal2016neural,chiunichols2016named}, we formulate the task as a sequence labeling problem. Given an input sequence consisting of $n$ tokens $(x_1, ..., x_n)$, the goal is to predict a sequence of labels $(y_1, ..., y_n)$, where $y_i$ is the label corresponding to token $x_i$. Table \ref{tab:tagger_label_set} describes the label set. We follow the BIO format: \texttt{B} denotes the beginning of a named entity, \texttt{I} denotes the continuation of a named entity, and \texttt{O} corresponds to tokens that are not part of any named entity.

\renewcommand{\arraystretch}{1.15}
\begin{table}[!t]
\small
\centering
\begin{tabular}{|l|p{4.5cm}|}
\hline
Tag & Description \\ \hline
\texttt{\{B,I\}-PER} & A named entity of a \textit{person}\\ \hline
\texttt{\{B,I\}-LOC} & A named entity of a \textit{location}\\ \hline
\texttt{\{B,I\}-GRP} & A named entity of a \textit{group}\\ \hline
\texttt{\{B,I\}-CORP} & A named entity of a \textit{corporation} \\ \hline
\texttt{\{B,I\}-PROD} & A named entity of a \textit{product} \\ \hline
\texttt{\{B,I\}-CW} & A named entity of a \textit{creative work} \\ \hline
\;\texttt{O} & Not a named entity \\\hline
\end{tabular}
\caption{The label set.}
\label{tab:tagger_label_set}
\end{table}

Figure \ref{fig:overview} shows a high-level overview of our Transformer-based model. Our model first forms a contextualized representation for each input token using a Transformer encoder \cite{devlinetal2019bert}. Let $\textbf{H} = (\textbf{h}_1, ..., \textbf{h}_n)$ be the output of the encoder where $\textbf{h}_i \in \mathbb{R}^{d}$. After that, we can predict the final label $y_i$ for each input token $x_i$:
\begin{equation}
\begin{split}
    \textbf{y}_i &= \text{softmax}(\text{FFNN}_\theta(\textbf{h}_i)) \\
    y_i &= \argmax_{j} \textbf{y}_{ij}
\end{split}
\end{equation}
where $\text{FFNN}_\theta$ is a trainable feedforward network. $\textbf{y}_i$ is the predicted probability distribution over the label set for the token $x_i$. The model is fine-tuned end-to-end via minimizing the typical cross-entropy loss.

Unlike many previous studies \cite{lampleetal2016neural,chiunichols2016named}, our model does not have a CRF layer \cite{Lafferty2001ConditionalRF}. A recent paper suggested that when using a pretrained Transformer language model for sequence labeling, adding a CRF layer may not improve the performance substantially \cite{Chen2019BERTFJ}.

\subsection{Data Augmentation} \label{sec:data_augmentation}
\begin{figure*}[ht!]
\centering
\includegraphics[width=0.85\textwidth]{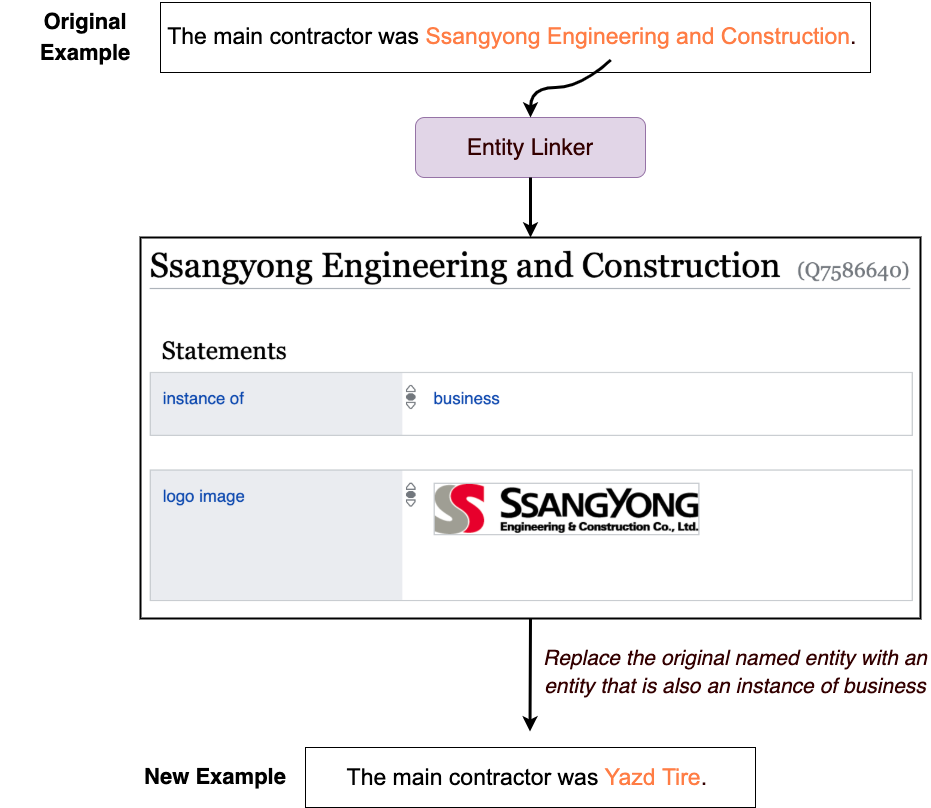}
\caption{Our data augmentation approach.}
\label{fig:data_augmentation}
\end{figure*}

To increase the size of the training set, we have also experimented with a simple data augmentation approach (Figure \ref{fig:data_augmentation}). For example, consider the sentence ``\textit{The main contractor was Ssangyong Engineering and Construction.}'', which is an example in the training set of the English track of MultiCoNER. In this case, ``\textit{Ssangyong Engineering and Construction}'' is a named mention referring to a Korean corporation. To create a new training example, we can replace the named mention with a different entity that is also a corporation.

More specifically, in this example, we first use an entity linker\footnote{\url{https://github.com/laituan245/EL-Dockers}} to link the named mention to its corresponding entity in Wikidata, a large-scale knowledge graph. From the found Wikidata page, we can extract all types of information about the entity. We can utilize these types of information to find a new entity that is different but highly similar to the original entity. For simplicity, in this work, we simply try to find a new entity of the same Wikidata type as the original entity. At the end, we will have a new example (e.g., ``\textit{The main contractor was Yazd Tire.}'').

\section{Results}
\label{sec:results}
\subsection{Data and Experimental Setup}
The learning rate is set to be 2e-5, and the batch size is 32. We experimented with different numbers of training epochs, 10 and 20. We use Huggingface’s Transformer library \cite{wolfetal2020transformers} to experiment with various Transformer language models:
\begin{itemize}
    \item \textbf{BERT}. \newcite{devlinetal2019bert} introduced a language representation model named BERT, which is pre-trained using two tasks: masked language modeling (MLM) and next sentence prediction (NSP). We used the large version of BERT (i.e., bert-large-uncased) in this work.
    \item \textbf{RoBERTa}. \newcite{Liu2019RoBERTaAR} proposed an improved recipe for training BERT models. The modifications include: (1) training the model longer, with bigger batches, over more data; (2) removing the NSP objective; (3) training on longer sequences; and (4) dynamically changing the masking pattern applied to the training data. We used the large version of RoBERTa (i.e., roberta-large) in this work.
    \item \textbf{ALBERT}. \newcite{Lan2020ALBERTAL} introduced ALBERT, a BERT-based model with two parameter reduction techniques: factorized embedding parameterization and cross-layer parameter sharing. We used the xxlarge version of ALBERT (i.e., albert-xxlarge-v2) in this work.
    
\end{itemize}
\renewcommand{\arraystretch}{1.75}
\begin{table*}[!ht]
\centering
\footnotesize
\begin{tabular}{|p{7cm}|p{7cm}|}
\hline
Original Example & Generated Example \\ \hline
\textcolor{blue}{the guardian} described the album 's release as one of the 50 key events  ... & \textcolor{red}{metro} described the album 's release as one of the 50 key events  ... \\ \hline
the game uses a battery packed \textcolor{blue}{random-access memory} in order to save progress . &  the game uses a battery packed \textcolor{red}{delay line memory} in order to save progress . \\ \hline
in the end the best placed rider was \textcolor{blue}{wilfried cretskens} who finished 61st . & in the end the best placed rider was \textcolor{red}{harald andersson} who finished 61st .\\ \hline
it was broadcast on the channel \textcolor{blue}{animal planet} , with episodes having aired between 2001 and 2003 . & it was broadcast on the channel \textcolor{red}{true4u} , with episodes having aired between 2001 and 2003 .  \\\hline
\end{tabular}
\caption{Some of the newly generated examples.}
\label{tab:generated_examples}
\end{table*}

\subsection{Results on the Development Set}

Table \ref{tab:dev_results} shows the overall results on the development set of the English track of MultiCoNER. We see that ALBERT-xxlarge trained with 20 epochs outperforms all other baseline models on the development set. As such, we use this model to generate predictions for the test set. The model achieved a macro F1 score of 72.50\% on the held-out test set. Note that the baseline models shown in Table \ref{tab:dev_results} are trained using only the original training set (without any data augmentation).

\begin{table}[t!]
\centering
\resizebox{\linewidth}{!}{%
\begin{tabular}{l|c|c|c}
 & Prec. & Recall & F1 \\ \hline
RoBERTa-large (10 epochs) & 85.63 & 87.82 & 86.68 \\\hline
BERT-large (10 epochs) & 86.02 & 88.34 & 87.14 \\\hline
ALBERT-xxlarge (10 epochs) & 86.81 & 88.7 & 87.7 \\\hline
ALBERT-xxlarge (20 epochs) & 86.47 & 89.49 & 87.91 
\end{tabular}%
}
\caption{Overall results on the development set. Macro scores (\%) are shown.}
\label{tab:dev_results}
\end{table}

\subsection{Analysis of the Data Augmentation Approach}
For each example in the training set, we used the data augmentation approach (Section \ref{sec:data_augmentation}) to generate a new example. Table \ref{tab:generated_examples} shows some of the newly generated examples.

We used all of the original and newly generated examples to train a new RoBERTa-large model (the number of epochs is 10). The model performs worst than the RoBERTa-large model trained with only the original examples. Nevertheless, we still believe the approach has a lot of potential, and we leave further exploration to future work.

\section{Conclusion} \label{sec:conclusion}
In future work, we plan to conduct a thorough error analysis and apply visualization techniques to understand our models better \cite{murugesan2019deepcompare}. In addition, as pretrained Transformer models are typically computationally expensive and have many parameters, we are also interested in reducing the computational complexity of our baseline models using compression techniques \cite{Sanh2019DistilBERTAD,lai2020simple,sunetal2020mobilebert}.


\bibliography{anthology,custom}
\bibliographystyle{acl_natbib}

\end{document}